\documentclass[journal,twoside,web]{ieeecolor}
\usepackage{tmi}
\usepackage{cite}
\usepackage{amsmath,amssymb,amsfonts}
\usepackage{algorithmic}
\usepackage{graphicx}
\usepackage{textcomp}
\usepackage{booktabs}
\usepackage{xcolor,colortbl}
\usepackage{multirow}
\newcommand{\Loss}{\mathcal{L}}
\usepackage{algorithm}
\usepackage{algorithmic}
\usepackage{xspace}
\usepackage{bm}
\usepackage{pifont}
\usepackage[outdir=./]{epstopdf}
\newcommand{\cmarkcolor}{\textcolor{teal}{\ding{51}}}%
\newcommand{\xmarkcolor}{\textcolor{red}{\ding{55}}}%
\newcommand{\cmark}{\ding{51}}%
\newcommand{\xmark}{\ding{55}}%
\definecolor{gray0}{gray}{0.95}
\definecolor{gray1}{gray}{0.9}
\definecolor{gray2}{gray}{0.85}

\makeatletter
\DeclareRobustCommand\onedot{\futurelet\@let@token\@onedot}
\def\@onedot{\ifx\@let@token.\else.\null\fi\xspace}
\def\etal{\emph{et al}\onedot}
\def\eg{\emph{e.g}\onedot} 
\def\ie{\emph{i.e}\onedot}

\def\wrt{\emph{w.r.t}\onedot}

\def\BibTeX{{\rm B\kern-.05em{\sc i\kern-.025em b}\kern-.08em
    T\kern-.1667em\lower.7ex\hbox{E}\kern-.125emX}}
\begin{document}
\title{Federated Learning with Privacy-Preserving Ensemble Attention Distillation}
\author{Xuan Gong, Liangchen Song, Rishi Vedula, Abhishek Sharma, Meng Zheng, Benjamin Planche, \\ Arun Innanje, Terrence Chen~\IEEEmembership{Senior Member, IEEE}, Junsong Yuan~\IEEEmembership{Fellow, IEEE},\\ David Doermann~\IEEEmembership{Fellow, IEEE}, Ziyan Wu~\IEEEmembership{Senior Member, IEEE}\thanks{X. Gong, L. Song, R. Vedula, J. Yuan and D. Doermann are with University at Buffalo, Buffalo NY, USA (\{xuangong, lsong8, rishisat, jsyuan, doermann\}@buffalo.edu). A. Sharma, M. Zheng, B. Planche, A Innanje, T. Chen and Z. Wu are with United Imaging Intelligence, Cambridge MA, USA (\{first.last\}@uii-ai.com). This paper is primarily based on the work done during X. Gong and L. Song's internships with United Imaging Intelligence.  
Corresponding author: Z. Wu.}
}

 

\maketitle

\begin{abstract}
Federated Learning (FL) is a machine learning paradigm where many local nodes collaboratively train a central model while keeping the training data decentralized.  This is particularly relevant for clinical applications since patient data are usually not allowed to be transferred out of medical facilities, leading to the need for FL. Existing FL methods typically share model parameters or employ co-distillation to address the issue of unbalanced data distribution. However, they also require numerous rounds of synchronized communication and, more importantly, suffer from a privacy leakage risk. We propose a privacy-preserving FL framework leveraging unlabeled public data for one-way offline knowledge distillation in this work. The central model is learned from local knowledge via ensemble attention distillation. Our technique uses \textcolor{black}{decentralized and} heterogeneous local \textcolor{black}{data} 
like existing FL approaches, but more importantly, it significantly reduces the risk of privacy leakage. We demonstrate that our method achieves very competitive performance with more robust privacy preservation based on extensive experiments on image classification, \textcolor{black}{segmentation}, and reconstruction tasks.
\end{abstract}


\vspace{-5pt}
\section{Introduction}
With increasing interest in topics such as edge computing \cite{li2018learning}, a new machine learning paradigm called federated learning (FL) is emerging. In FL, one does not necessarily need all data samples to reside in \textcolor{black}{one specific} 
``local" or edge-compute node. Instead, it relies on model fusion/distillation techniques to train a single centralized model in a distributed, decentralized fashion.  

Several vital challenges make FL markedly different from typical distributed learning.  First, privacy is always the key concern, \textcolor{black}{\wrt,} protecting local data. Second, communication is a critical bottleneck, as the central training can \textcolor{black}{easily} get disrupted by network communication issues.  Third,  individual local data distributions can differ substantially as distributed data centers tend to collect data in different settings. This inherent heterogeneity can manifest itself in various ways: \textcolor{black}{various} 
\textcolor{black}{data size or} domain distributions, different local model architectures, or simply the diversity in knowledge across all local models.

\begin{figure}
\begin{center}
\includegraphics[width=1\linewidth]{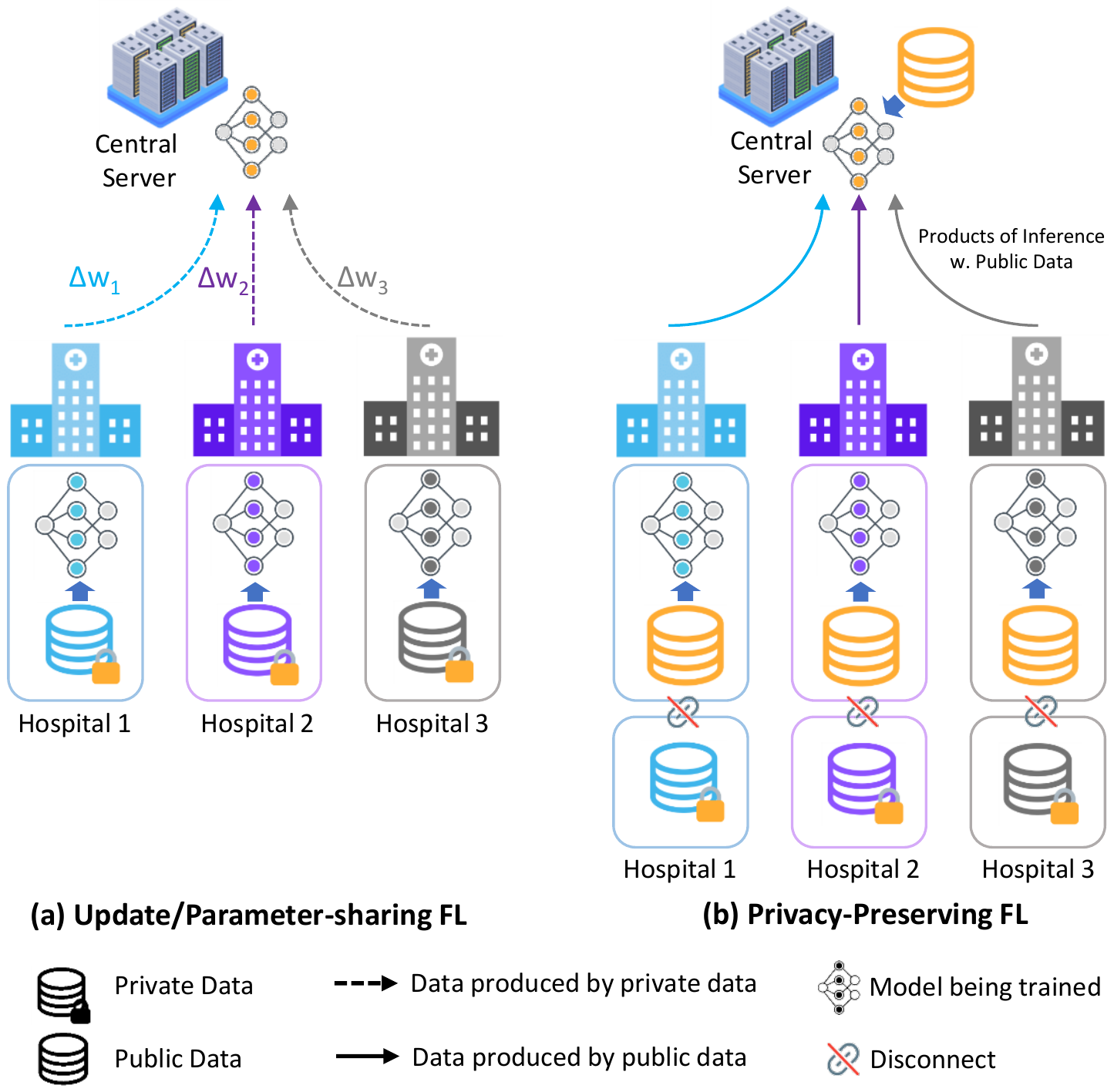}
\caption{Unlike traditional FL approaches (a) which exchange gradients or weights directly with local nodes, the proposed privacy-preserving FL framework (b) relies only on the exchange of products of inference with non-proprietary public data. 
}
\label{fig:teaser}
\vspace{-10pt}
\end{center}
\end{figure}

Mainstreamed federated learning methods teach the central model using a recursive exchange of parameters/gradients between the central and local nodes~\cite{mcmahan2017communication, smith2017federated, hsu2019measuring, li2018federated, zhao2018federated, wang2020federated, karimireddy2019scaffold}. 
Recent applications of FL in medical imaging also share similar features. Sheller \etal \cite{sheller2018multi} share local model gradients for multi-institution collaboration, and Yang \etal \cite{yang2021federated} employ a similar approach for the task of semi-supervised COVID segmentation. Typically, such methods involve each local model sharing its gradients with a central server after each round of local training on its local data.  The central server then aggregates the local model parameters \cite{wang2020federated,li2019fair,hsu2020federated}. Each local node then updates its local model with the latest global aggregation update, and this process repeats till convergence. These parameter-based communication methods have many known security weaknesses and are limited to models with homogeneous architectures. While some methods have shown hope of protecting against data leakage in medical imaging \cite{li2019privacy,li2020multi}, some recent works~\cite{zhu2019deep, geiping2020inverting} demonstrated that local private data could be \textcolor{black}{fully} recovered from publicly-shared gradients, further highlighting the associated privacy-related risks in general and in medical applications in particular. 

Another approach to fusing local models into a single central model is to employ distillation~\cite{li2019fedmd,chang2019cronus}, where the central model is learned as a student with multiple local models as teachers. 
Distillation-based methods train the central model with aggregated locally-computed logits~\cite{jeong2018communication, chang2019cronus, lin2020ensemble}, \textcolor{black}{therefore} eliminating the requirement of identical architectures. 
Despite the known bottleneck of communication in FL, existing distillation-based FL \textcolor{black}{methods} optimize the central and local models jointly by synchronizing local inferred predictions, requiring a high degree of synchronization and numerous rounds of communication. Moreover, these techniques either assume that both the public and private data are sampled from the same underlying distribution \textcolor{black}{\cite{li2019fedmd}}, or iteratively exchange local parameters beyond the products on public data 
\textcolor{black}{\cite{lin2020ensemble},} both invariably exposing private data to attack. 
In addition to the issues of network communication and data security issue\textcolor{black}{s} discussed above, 
these existing methods mainly rely on distilling the final predictions, \eg, logits, and completely ignore the structure knowledge that leads to these final predictions. 

In contrast to previous FL methods that incrementally train the local models, update them synchronously, and exclusively share parameters or distill logits online, this paper ensembles stale local information with both logits and feature-level knowledge using a privacy-preserving, offline, federated distillation method under the heterogeneous FL setting, as illustrated in Figure~\ref{fig:teaser}.  
\textcolor{black}{To protect the privacy of local data,} we distill the unlabeled, non-sensitive, cross-domain public data output without exchanging local model parameters or gradients. The proposed distillation is one-way, \ie, from local nodes to a central server, with the local training remaining asynchronous and independent. 
Our key insight is that training the local models to completion allows us to  mine and ensemble local models with structural knowledge to capture \textcolor{black}{the internal} expertise.  To coordinate local knowledge in the FL heterogeneity framework, we propose federated attention distillation (FedAD) to fully exploit the consensus and diversity of attention maps across local models.  We demonstrate the effectiveness and efficiency of our method
via experiments with chest x-ray and brain tumor MRI datasets on image classification, \textcolor{black}{segmentation}, and reconstruction tasks. Our contributions are summarized as follows:
\begin{itemize}
\item We propose an offline federated distillation framework to explicitly preserve local data privacy by only distilling model outputs on unlabeled, non-sensitive public data.

\item To deal with the heterogeneity in federated learning scenarios, we distill structure knowledge via novel attention-bound constraints for local ensembles with a trade-off between local consensus and diversity. 

\item Our federated distillation pipeline is model-agnostic and highly flexible without any requirement \wrt online synchronization during communication.

\item We \textcolor{black}{empirically} show that our method can be extended to typical medical tasks such as image reconstruction. We simultaneously achieve competitive performance with a more robust privacy-preserving guarantee and superior communication efficiency.
\end{itemize}

This paper is an extended version of our previous work \cite{Gong_2021_ICCV}. In particular, (a) we generalize our bound constraint from top-down attention generated by Grad-CAM \cite{selvaraju2017grad} to self-attention \cite{wang2018non}. (b) we extend the applications beyond classification or segmentation-related tasks to more general medical tasks, \ie, image reconstruction. (c) we theoretically analyze the \textcolor{black}{privacy guarantee of our method} (performance bound of the central model), where private data across local nodes, and public data used in knowledge distillation, are from different domains. (d) we provide comprehensive empirical studies on chest-x-ray image classification,  brain tumor image \textcolor{black}{segmentation} and reconstruction tasks on cross-domain federated distillation. The experiment settings \textcolor{black}{simulate the real in-the-wild FL scenarios, including} local data across different \textcolor{black}{institutions, local data with various sizes}, public data in different domains with local data, and public data with modalities different from local ones.


\vspace{-5pt}
\section{Related Work}
\label{sec:related}
\subsection{Federated Learning}
\subsubsection{Parameter-based FL}
In parameter-based FL methods, each local model shares its parameters/gradients with the central server after \textcolor{black}{each} round of local training on its local data. The central server aggregates them by averaging~\cite{mcmahan2017communication}. Then the aggregated results are shared with the local nodes, updating their corresponding local model before proceeding with the next training round. This process is repeated until the stopping criterion is met. A variety of extensions of FedAvg~\cite{mcmahan2017communication, wang2020federated, li2019fair, hsu2020federated} employ improved aggregation schemes, such as adding momentum~\cite{hsu2019measuring} and local weighting~\cite{li2019fair, hsu2020federated}. 
Local weighting schemes have also been investigated based on client loss~\cite{li2019fair} and client data size~\cite{hsu2020federated}.  FedMA~\cite{wang2020federated} aggregates local parameters layer-wise by matching and averaging hidden elements. FedProx~\cite{li2018federated} incorporates a proximal term to restrict local updates close to the global model. SCAFFOLD~\cite{karimireddy2019scaffold} introduces control variations to correct the local updates.

\begin{figure}
\begin{center}
\includegraphics[width=1\linewidth]{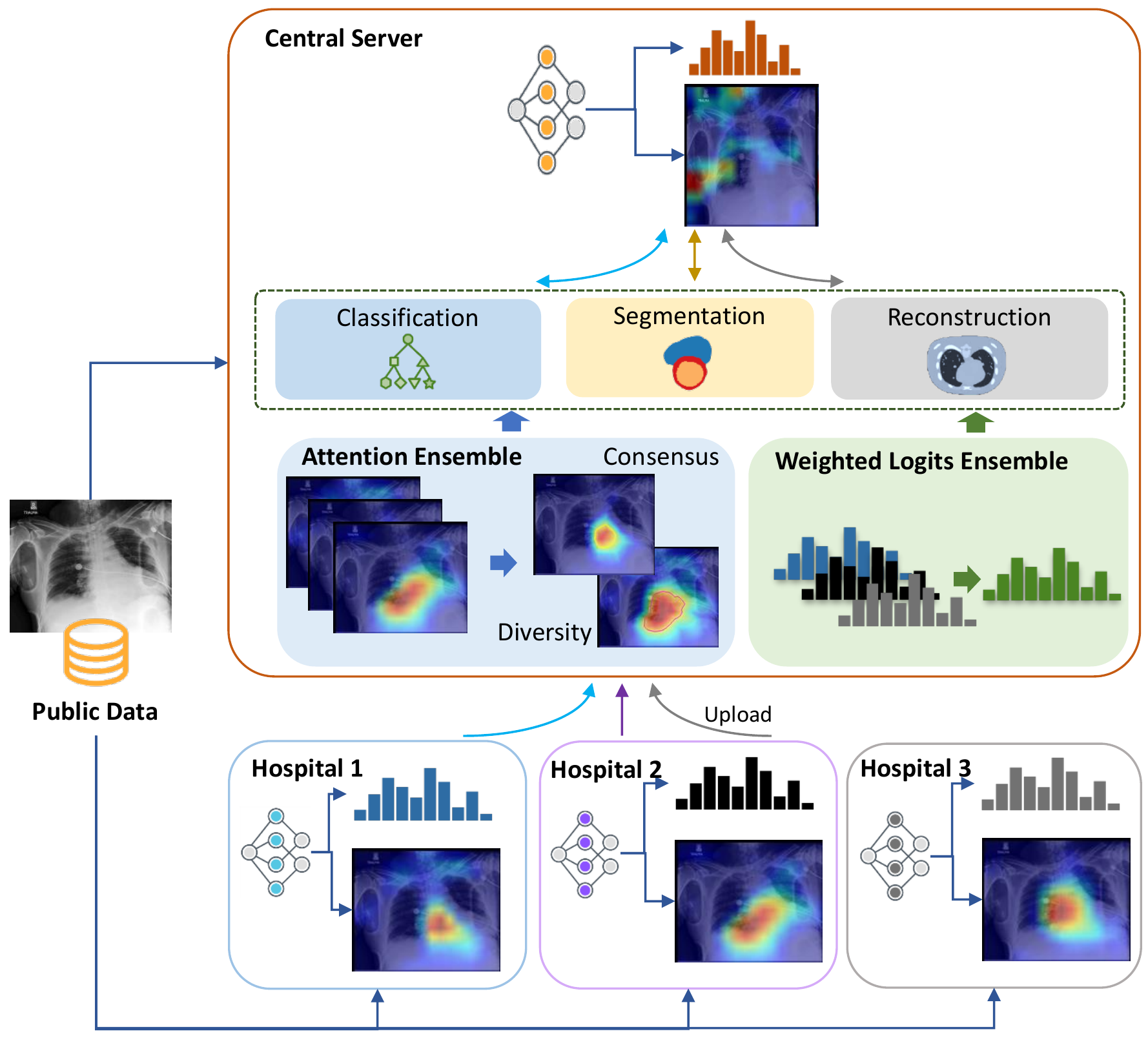}
\caption{Overall pipeline of the proposed FedAD framework. }
\label{fig:pipeline}
\vspace{-10pt}
\end{center}
\end{figure}

\subsubsection{Distillation-based FL}
Federated distillation methods exchange model output rather than model parameters.
Solutions that produce central models by distilling knowledge from private data incur concerns \wrt the leaking of private local data~\cite{zhou2020distilled, shin2020xor}. 
In contrast, some works~\cite{jeong2018communication, li2019fedmd, chang2019cronus} distill the output of public data.
FedMD~\cite{li2019fedmd} divides local training and joint distillation into two phases by adding labeled public data. 
Each local model is fully trained with public and private data in the first phase. In the collaboration phase, the central and local models are jointly optimized to reach a consensus via the distillation of predictions on the public data. Cronous~\cite{chang2019cronus} implements a similar two-phase system, but the local initialization only utilizes private data since the public data is unlabeled.  
Although model-agnostic, these methods select public data based on prior knowledge \wrt private data. 
The recently proposed FedDF~\cite{lin2020ensemble} makes it more robust to distillation data selection, but it still has privacy issues due to the iterative exchange of models over hundreds of rounds. 
The above-mentioned exclusively require many rounds of back-and-forth communication, leading to bandwidth bottlenecks and other inefficiencies.
For communication efficiency, Guha \etal \cite{guha2019one} proposes a preliminary investigation into offline federated learning where all local models can be
trained independently without inter-institutional communication. It then distills through the averaging of predictions on unlabeled data. 
Another similar work is PATE \cite{papernot2016semi}, where the central model is trained with hard pseudo-labels voted by local models rather than the soft prediction we employed.

\subsection{Knowledge Transfer}
\subsubsection{Knowledge Ensemble}
Model ensemble~\cite{zhou2002ensembling} 
is a popular regularization technique that is more robust and generalizable than individual models. 
Classical ensemble approaches such as bagging~\cite{caruana2004ensemble}, boosting~\cite{avnimelech1999boosted}, and Bayesian~\cite{bishop2012bayesian} are exploited to assemble partial knowledge for better prediction as a  mixture of experts \cite{jacobs1991adaptive}. With the success of knowledge transfer~\cite{hinton2015distilling}, recent advancements in ensemble networks are dominated by the student-teacher learning paradigm~\cite{shazeer2017outrageously}. Ensemble learning aggregates the knowledge of multiple teachers before distilling the knowledge into the student network. Supervised ensemble learning is dominated by gate learning to design the weight for aggregation~\cite{shazeer2017outrageously, asif2019ensemble, xiang2020learning}. In semi-supervised and self-supervised scenarios, \cite{wu2019distilled} and \cite{you2017learning} exploit the relative similarity between samples. Furthermore, co-distillation extends one-way transfer to bidirectional collaborative learning~\cite{song2018collaborative, zhu2018knowledge, guo2020online}. 

\subsubsection{Structure Knowledge}
Beyond the use of softened labels for distillation~\cite{hinton2015distilling}, many improvements have been developed to transfer structure knowledge, such as intermediate representations~\cite{romero2014fitnets}. Attention transfer relaxes feature-level knowledge to attention maps either from bottom-up activation~\cite{zagoruyko2017paying} or top-down gradients~\cite{dhar2019learning}. Yim \etal~\cite{yim2017gift} utilizes the Gram matrix as a flow of solution procedure for distillation. Other works match features such as maximum mean discrepancy~\cite{huang2017like} or mutual information~\cite{ passalis2020heterogeneous}. 
More recent attempts explore structured knowledge ensemble for knowledge distillation with labeled distillation data. FEED~\cite{park2020feed} can be seen as an extension of self-distillation~\cite{furlanello2018born} that accumulates and assembles feature-level knowledge to train itself recursively. Knowledge flow~\cite{liu2019knowledge} enhances the student by adding transformed and scaled intermediate representations from teacher models. In contrast, our feature-level ensemble method is label-free, model agnostic, and is used for heterogeneous federated distillation scenarios.

\vspace{-5pt}
\section{Methodology}
\label{sec:method}

\subsection{Problem Definition}
In FL scenarios, we consider $K$ local nodes, each hosting a private, local dataset $\mathcal{D}_k=\{(x_k^i, y_k^i)|i=1,\ldots,|\mathcal{D}_k|\}$, where $x_k^i$ and $y_k^i$ are the $i$-th paired data sample from the $k$-th local node. 
The public dataset $\mathcal{D}_0=\{x_0^i|i=1,\ldots,|\mathcal{D}_0|\}$ can be either labeled or unlabeled and is accessible by all local nodes.

As illustrated in Figure~\ref{fig:pipeline}, 
in the first stage, the model at each local node $k$ is initialized by training over its own local, private data $\mathcal{D}_k$. Let $\theta_k$ denote the model parameters after this initial training. Please note that the proposed distillation-based approach is agnostic to neural network architecture. Hence, each local node can specify its unique architecture suited to the particular distribution of its local data.
In the second stage, we ignore all private datasets and freeze the local models; hence reducing the risk of any data leakage through the exposure of the local models or data.
Instead, the public dataset $\mathcal{D}_0$ hosted on the server and deployed at each local node is used for a one-way knowledge distillation procedure, from locals to the server. Each fully-trained local model $\theta_k$ and the server-based central model $\theta_\mathrm{s}$ constitute a teacher-student knowledge transfer setup%
. Overall, we consider an ensemble of multiple teachers, one at each local node, which only communicate products inferred on public data $\mathcal{D}_0$ to the server-based student.

\subsection{Ensemble and Distillation}
\subsubsection{Conventional Ensemble Distillation}
Let $z^{c}_k=f(x_0,\theta_k, c)$  be the logits of a public data sample $x_0$ corresponding to class $c \in \mathcal{C}_k$, produced by the model at local node $k$, and the output of the central model be $\tilde {z}^c=f(x_0,\theta_\mathrm{s}, c)$, where $c \in \{1,\ldots,C\}$.
The conventional ensemble $\hat{z}^{c}= \frac{1}{K}\sum_{k = 1}^K z^{c}_k$ takes an average of all teachers' logits and then employ activation $\sigma(\cdot,\cdot)$ using softmax to represent the probabilities that the sample belongs to class $c$ for:
\begin{equation}\label{eq2}
   \sigma(z^c,\tau) = \frac{exp(z^c/\tau)}{\sum_c{exp(z^c/\tau)}}
\end{equation}
where $\tau$ is a temperature parameter. 
Taking the teachers' ensembled soft labels as  $\sigma(\hat{z}^{c},\tau)$ and the student soft label as $\sigma(\tilde{z}^{c},\tau)$ , conventional knowledge  distillation employs the Kullback-Leibler divergence to update the student model:
\begin{equation}\label{eq1}
    \Loss =  \sum_c \sigma(\hat{z}^{c},\tau) \mathrm{log}{\frac{ \sigma(\hat{z}^{c},\tau)}{ \sigma(\tilde{z}^{c},\tau)}}.
\end{equation}
Hinton \etal \cite{hinton2015distilling} has shown that minimizing Eq.~\ref{eq1} with a high $\tau$ is equivalent to minimizing the $\ell_2$ error between the logits of teacher and student, thereby relating cross-entropy minimization to matching logits. 

\subsubsection{Importance of Weighted Ensemble Distillation}
Let the global data distribution $p_0(x,y)$ of image $x$ and label $y$ be the target, while the local private data distribution can be indicated as $p_k(x,y)$. 
Due to the imbalance of data distribution among locals, we come to the bias ratio of local prediction :
\begin{equation}
    \hat{p}_k(x,y)= \frac{p_k(x,y)}{p_0(x,y)} = \frac{p_k(y)p_k(x|y)}{p_0(y)p_0(x|y)} \thickapprox  \frac{p_k(y)}{p_0(y)},
\end{equation}
where we assume $p_k(x|y) \thickapprox p_0(x|y)$ as the local difference on conditional probability distribution $p_k(x|y)$ is minor.

To consider this aspect, during the ensemble, we introduce an importance weight $\omega$ for each local node to reflect the distribution of local private data that its model was initially trained with:
\begin{equation}
\label{eqweight2}
\hat{z}^{c} = \sum_{k} {\omega_k^c z^c_k}, 
~\omega_k^c = \frac{N_{k}^c}{\sum_k {N_{k}^c}},
\end{equation}
where the importance weight $\omega^c$ is class-specific, which means the number of samples labeled as class $c$ for training the model of local node $k$: $N_{k}^c= \sum_{i=1}^{|\mathcal{D}_k|}({y}_k^i(c)= 1 )$. It reflects the distribution of local private data corresponding to each particular class $c$. Without loss of generality, we denote the local model output as $z_{k}=f(x_0,\theta_k)$ and central model output as $\tilde{z}=f(x_0,\theta_\mathrm{s})$. Specifically, for the classification task, we have $z_{k}=[z_{k}^1, ..., z_{k}^C]$ and $\tilde{z}=[\tilde{z}^1, ..., \tilde{z}^C]$ for local model and central model respectively.
Following the aforementioned $\ell_2$ observation from \cite{hinton2015distilling}, we consider the case of $\tau \rightarrow \infty$, hence expressing the logit loss as:
\begin{equation} \label{eqlogits}
    \Loss_\text{w}({\tilde z}, {\hat{z}}) = \| {\tilde z} - {\hat{z}}\|.
\end{equation}

\subsubsection{Attention Bounded Ensemble Distillation}
The above-mentioned distillation essentially captures the divergence between the final output from the teacher and student models. However, it provides little insight into the underlying structure knowledge or reasoning of the teacher models, which can be complementary and important to the final output, especially in the FL scenario with its high degree of heterogeneity in local data sources.
Although intuitive, it challenges transferring structural and more comprehensive knowledge. 
Structural knowledge, such as intermediate feature representations, suffers from a high bandwidth burden and, in most cases, relies on an identical network architecture among central (student) and all the local (teacher) models.  
Therefore we turn to more concise attention interpretations to transfer knowledge in a more efficient (as opposed to full feature tensors) and effective (as opposed to output vectors only) way without risking privacy leakage or posting additional communication/architecture requirements. 

Specifically, we propose a bounded constraint for attention ensemble  distillation to achieve consensus while maintaining the local node's' inherent diversity. Let $\bm{A}_k \in \mathbb{R}^{HW}$ be the attention map produced by the $k$-th local model, where $H$ and $W$ represent the 2D size of the attention map. Given the set of local attention maps  $\mathcal{A} = \{\bm{A}_k|k = 1, ..., K\}$, we take the spatial-wise minimum among them 
as the attention consensus $\bm{I}$, and take the spatial-wise maximum to represent the attention diversity $\bm{U}$ among $\mathcal{A}$:
\begin{equation} \label{eq:attiu}
    I^{hw} = \min \limits_{k} A_k^{hw}, U^{hw} = \max \limits_{k}A_k^{hw},
\end{equation}
where $h=1,...,H$ and $w=1,...,W$. $\bm{I}$ denotes a consensus on the high-response region, among all the local attention maps, that have a high probability to be the real attention. While $\bm{U}$ considers all the high-response regions among the local attention maps, it also preserves diversity of ``expertise" among the local models. 

For simplicity, we denote $\tilde{\bm{A}}$ as the attention map generated by the central model. Considering the attention consensus $\bm{I}$ as a lower bound, we enforce the response in $\tilde{\bm{A}}$ to explicitly activate at the region of consensus achieved by all locals:
\begin{equation}\label{eq:lbound}
    \Loss_\text{low}(\tilde{\bm{A}}, \bm{I}) = -\frac{1}{HW}\frac{\sum_{h,w} I^{hw} \cdot T(\tilde{A}^{hw}) }{\sum_{h,w} I^{hw}}.
\end{equation}
$T(\cdot)$ is a soft-masking operation based on sigmoid~\cite{li2018tell}:
\begin{equation}
T({A}) = \frac{1}{1+exp(-\rho({A}-b))}.
\end{equation}

Considering all the high-response regions among locals $\bm{U}$ as the upper bound, we enforce the response of $\tilde{\bm{A}}$ to be explicitly lower than that of $\bm{U}$:
\begin{equation}\label{eq:ubound}
    \Loss_\text{up}(\tilde{\bm{A}}, \bm{U}) = -\frac{1}{HW}\frac{\sum_{h,w} \tilde{A}^{hw} \cdot T(U^{hw}) }{\sum_{h,w} \tilde{A}^{hw}},
\end{equation}
The intuition here is that we seek each high-response pixel in $\tilde{\bm{A}}$ to have support from at least one local model to consider model diversity. 

Compared to the naive distillation that enforces precisely the same attention strength as one \textcolor{black}{aggregated} attention map of diverse local nodes,  our designed attention bound constraint is a relaxed version, \ie, tolerating incorrect/biased local attention maps. Its high robustness to outliers enables it to handle the inherent heterogeneity among locals more efficiently.

\begin{figure*}
\vspace{-15pt}
\begin{center}
\includegraphics[width=1\linewidth,trim={0 163pt 0 0},clip]{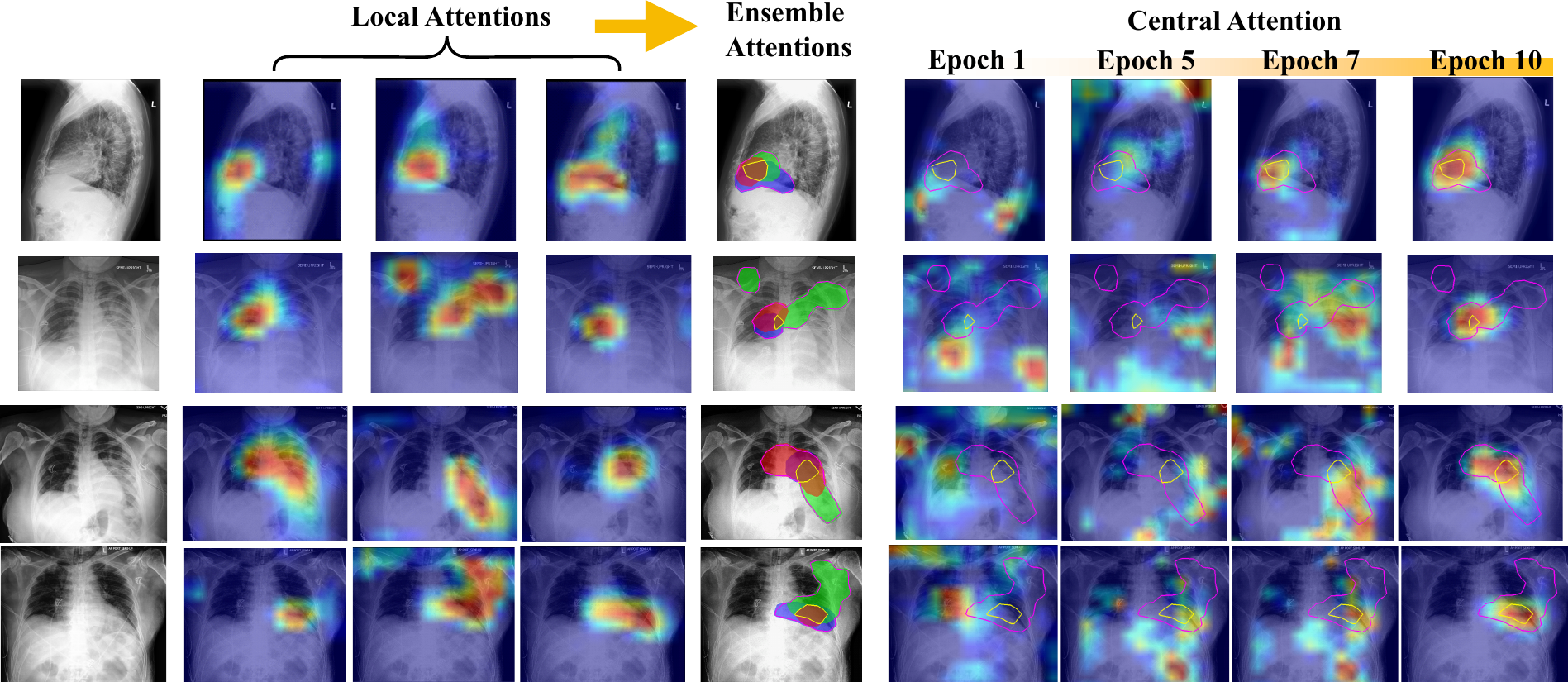}
\caption{Illustration of how ensemble attention can effectively guide the central model to focus on the correct region. In ensemble attention visualization, we threshold the activation at 0.5 and color the activated area red, green, and blue for the three locals, respectively. We contour the boundary of attention consensus/diversity \textcolor{black}{region} with yellow/black.
}
\label{fig:cxratt}
\vspace{-10pt}
\end{center}
\end{figure*}

\subsection{Application to different tasks}
\subsubsection{Image Classification / Segmentation}
For the classification task, the heterogeneity of FL mainly comes from its inability to cope with the more general scenario of local nodes not sharing the same target classes. We note that Eq. \ref{eqweight2} can deal with this issue efficiently.

We employ top-down attention generated with  Grad-CAM~\cite{selvaraju2017grad} to provide location cues for class activation reasoning.  
Feeding an image to the model obtains a raw score $z^c$ (before the activation layer) for each class $c$. The gradient of score $z^c$ is computed with respect to the feature maps in a convolutional layer $[\bm{F}_1, \bm{F}_2, ..., \bm{F}_J]$, where $\bm{F}_j \in \mathbb{R}^{H \times W}$, $J$ is the channel size, $H$ and $W$ indicates size of a 2D feature. These gradients can be globally averaged to obtain the neuron importance $\beta_j^c$ corresponding to $\bm{F}_j$:
\begin{equation}
    \beta_j^c = \frac{1}{HW} \sum_h \sum_w \frac{\partial z^c}{\partial F_j^{hw}}.
\end{equation}
All $\bm{F}_j$ weighted by $\beta_j^c$ are combined and activated with ReLU to get the a class-specific attention map $\bm{A}^c \in \mathbb{R}^{H \times W}$: 
\begin{equation}
\label{eq:gradcam}
    \bm{A}^c = \text{ReLU}(\sum_{j}{\beta_j^c \cdot \bm{F}_j}).
\end{equation}
We then normalize the attention maps to have all values lie between 0 and 1: $A_{hw}^c = \frac{A_{hw}^c }{ \max_{hw} A_{hw}^c}$.
When employing the attention-bound loss, Eq.\ref{eq:lbound} and  Eq.\ref{eq:ubound}, the class-specific attention maps are taken independently, and thus the overall loss function for classification can be written as:
\begin{equation}
\label{eq:lossclassify}
   \Loss_\text{cls} = \frac{1}{C} \sum_c \Loss_\text{w}(\tilde{z}^c, \hat{z}^c) +  \Loss_\text{low}(\tilde{\bm{A}}^c, \bm{I}^c) + \Loss_\text{up}(\tilde{\bm{A}}^c, \bm{U}^c) .
\end{equation}
Segmentation can be seen as pixelwise/voxelwise classification task while differing from the above mainly in two aspects: 1) the model's prediction $\bm{z}^c \in \mathbb{R}^{HW}$ is the same shape as its input (denoted as 2D here for simplicity); 2) we directly employ $\bm{z}^c$ for the attention  $\bm{A}^c=\sigma(\bm{z}^c, \tau)$ relating to the activation bound constraint.

\subsubsection{Image Reconstruction}
For image reconstruction tasks, we rewrite the weighted ensemble as:
\begin{equation}\label{eqweight}
\bm{\hat{z}}= \sum_{k} {\omega_k \cdot \bm{z}_k}, 
~\omega_k =  \frac{|\mathcal{D}_k|}{\sum_k |\mathcal{D}_k|},
\end{equation}
where $|\mathcal{D}_k|$ denotes the number of samples used to train the model at local node $k$, and model output ${\bm{z}}$ is with 2D image size. 
We employ a non-local self-attention module \cite{wang2018non} to capture spatial-wise dependencies  of 2D features. For one batch, given the feature maps $\bm{F}$ with size $J\times H \times W$, we reshape $\bm{F}$ to $\bar{\bm{F}}$ (with size $J\times HW$) and then calculate the spatial-wise similarity $\bm{S} \in \mathbb{R}^{HW \times HW}$ via dot product (matrix multiplication): $\bm{S} = \bar{\bm{F}}^\text{T} \cdot \bar{\bm{F}}$.
Then $\bm{S}$ is normalized into spatial-wise attention $\bm{A}$ using softmax along the first dimension:
\begin{equation}
\label{eq:nonlocalatt}
  A^{hw} = \frac{exp(S^{hw})}{\sum_{h=1}^{HW}exp(S^{hw})}.
\end{equation}
The normalized similarity is used to enhance the features $\bm{F}$:
\begin{equation}
\label{eq:nlocal}
  \bm{F}= \bm{F} + \text{Reshape} (({\bm{A}} \cdot \bar{\bm{F}}^\text{T})^\text{T}),
\end{equation}
where $\text{Reshape}(\cdot)$ is to reshape the size of $J\times HW$ to $J\times H \times W$.
The overall loss function for image reconstruction can be written as:
\begin{equation}
\label{eq:lossrecon}
   \Loss_\text{recon} =  \Loss_\text{w}(\tilde{\bm{z}}, \hat{\bm{z}}) +  \Loss_\text{low}(\tilde{\bm{A}}, \bm{I}) + \Loss_\text{up}(\tilde{\bm{A}}, \bm{U}).
\end{equation}

The overall process is explained in Algorithm~\ref{alg}.
\begin{algorithm}[t]
    \caption{FedAD on {\color{cyan} classification}/ {\color{blue} reconstruction} }
    \label{alg}
 \begin{algorithmic}
    \STATE {\bfseries Input:} Labeled private dataset $\{\mathcal{D}_k\}$, unlabeled public data $\mathcal{D}_0$, central model $\theta_\mathrm{s}$, local models $\{\theta_k\}$, $T$ distillation steps.
    \STATE {\bfseries Local Training:} Train each local model $\theta_k$  with $\mathcal{D}_k$
    \FOR{each distillation step $t=1,...,T$}
    \STATE $\bm{x}_0$ $\leftarrow$ a batch of public data from $\mathcal{D}_0$ 
    \FOR {$k =1 ,..., K$ }
    \STATE $\bm{z}_k$, $\bm{A}_k$ $\leftarrow f(\bm{x}_0, \theta_k)$ \hspace*{6em} $\triangleright$ {\color{cyan} Eq.~\ref{eq:gradcam}}/ {\color{blue} Eq.~\ref{eq:nonlocalatt} }
    \ENDFOR
    \STATE $\hat{\bm z} \leftarrow$ ensemble $\{\bm{z}_k\}$  \hspace*{6.9em}   $\triangleright$ {\color{cyan} Eq.~\ref{eqweight2}}/ {\color{blue} Eq.~\ref{eqweight} }
    \STATE ${\bm I}, {\bm U} \leftarrow$ ensemble $\{\bm{A}_k\}$  \hspace*{5.3em} $\triangleright$ { Eq.~\ref{eq:attiu}}
    
    \STATE $\tilde{\bm z}, \tilde{\bm A}  \leftarrow f(\bm{x}_0, \theta_\mathrm{s})$ \hspace*{8.2em} $\triangleright$ {\color{cyan} Eq.~\ref{eq:gradcam}}/ {\color{blue} Eq.~\ref{eq:nonlocalatt} }
    \STATE Update: $\theta_\mathrm{s} \leftarrow {\theta_\mathrm{s}} - \nabla_{\theta_\mathrm{s}} \Loss$ \hspace*{5.3em} $\triangleright$ {\color{cyan} Eq.~\ref{eq:lossclassify}}/ {\color{blue} Eq.~\ref{eq:lossrecon} }
    \ENDFOR
\end{algorithmic}
\end{algorithm}

\begin{table*}
\centering
\vspace{-10pt}
\caption{Results on CXR14 and CheXpert with in/cross-domain local nodes ($K_d$=3, $\alpha=1$). Models are tested on NIH CXR14 (12 classes) and CheXpert (8 classes).  ``Centralized" denotes the result of centralized training with all samples from one dataset. ``Standalone" denotes the average performance of all distributed local models. ``E.Cardio" abbreviates ``Enlarged Cardiomediastinum". When training and testing samples are from different datasets, only the mean AUC on the six overlapping classes are listed (mAUC$^{\textit{overlap}}$). `mAUC$^{\textit{total}}$' denotes the mean AUC on all classes of the two test sets, which are 12 and 8 for CXR14 and CheXpert respectively.} 
\resizebox{\textwidth}{!}{
\begin{tabular}{c|c>{\columncolor[gray]{0.9}}c|c>{\columncolor[gray]{0.9}}c|c>{\columncolor[gray]{0.9}}c|c>{\columncolor[gray]{0.9}}c|c>{\columncolor[gray]{0.9}}c|c>{\columncolor[gray]{0.9}}c|c>{\columncolor[gray]{0.9}}c}
\toprule
 & \multicolumn{6}{c|}{NIH CXR14} & \multicolumn{6}{c|}{CheXpert} & \multicolumn{2}{c}{Cross-domain} \\\cmidrule(lr){2-15}
&\multicolumn{2}{c|}{Centralized} 
&\multicolumn{2}{|c|}{Standalone} 
&\multicolumn{2}{|c|}{FedAD} 
&\multicolumn{2}{|c|}{Centralized} 
&\multicolumn{2}{|c|}{Standalone} 
&\multicolumn{2}{|c|}{FedAD} 
&\multicolumn{2}{|c}{FedAD} \\
Pathology &\textit{test}$^{\text{\textit{CXR}}}$ &\textit{test}$^{\text{\textit{CheXpert}}}$   &\textit{test}$^{\text{\textit{CXR}}}$ &\textit{test}$^{\text{\textit{CheXpert}}}$  &\textit{test}$^{\text{\textit{CXR}}}$ &\textit{test}$^{\text{\textit{CheXpert}}}$  &\textit{test}$^{\text{\textit{CXR}}}$ &\textit{test}$^{\text{\textit{CheXpert}}}$  &\textit{test}$^{\text{\textit{CXR}}}$ &\textit{test}$^{\text{\textit{CheXpert}}}$  &\textit{test}$^{\text{\textit{CXR}}}$ &\textit{test}$^{\text{\textit{CheXpert}}}$  &\textit{test}$^{\text{\textit{CXR}}}$ &\textit{test}$^{\text{\textit{CheXpert}}}$\\ \midrule
Cardiomegaly &86.18 & 82.72 & 81.34 & 64.73 & 78.4 & 52.22 & 84.18 & 77.94 & 71.81 & 62 & 67.91 & 72.14 & 82.35 & 68.91 \\
Emphysema & 89.21 & - & 84.29 & - & 81.75 & - & - & - & - & - & - & - & 82.13 & - \\
Hernia & 85.39 & - & 83.63 & - & 84.57 & - & - & - & - & - & - & - & 83.62 & - \\
Infiltration &70.97 & - & 63.17 & - & 69.61 & - & - & - & - & - & - & - & 69.49 & - \\
Mass &78.86 & - & 73.47 & - & 74.5 & - & - & - & - & - & - & - & 74.91 & - \\
Nodule &77.38 & - & 69.23 & - & 70.56 & - & - & - & - & - & - & - & 69.99 & - \\
Atelectasis &76.57 & 88.17 & 72.27 & 71.85 & 72.6 & 74.42 & 72.16 & 76.64 & 66.24 & 76.12 & 57.72 & 84.03 & 71.93 & 84.41 \\
Pneumothorax &85.56 & 78.86 & 81.18 & 73.64 & 79.47 & 78.04 & 84.9 & 84.95 & 74.35 & 64.7 & 61.78 & 64.78 & 80.82 & 72.76 \\ 
Pneumonia &70.77 & 75.57 & 68.72 & 72.07 & 68.26 & 79.83 & 70.54 & 84.91 & 65.06 & 75.9 & 64.98 & 90.32 & 68.4 & 93.27 \\ 
Fibrosis &80.35 & - & 74.12 & - & 72.31 & - & - & - & - & - & - & - & 72.39 & - \\ 
Edema &82.96 & 85.84 & 80.6 & 74.63 & 80.4 & 83.37 & 79.41 & 88.9 & 77.33 & 84.17 & 75.32 & 83.82 & 80.39 & 80.95 \\ 
Consolidation & 72.6 & 85.29 & 68.96 & 75.68 & 69.57 & 89.03 & 70.93 & 92.56 & 63.8 & 83.85 & 67.01 & 92.56 & 70.08 & 94.33 \\
E.Cardio. &- & - & - & - & - & - & - & 61.21 & - & 57.72 & - & 83.66 & - & 79.64 \\ 
Lung Opacity &- & - & - & - & - & - & - & 93.5 & - & 84.89 & - & 89.65 & - & 86.95 \\\midrule
\# class &12 &6 &12 &6 &12 &6 &6 &8 &6 &8 &6 &8 &12 &8 \\ \hline
mAUC$^{\textit{overlap}}$ &\textbf{79.11} &82.74 &74.51 &72.1 &74.78 &76.15 &77.02 &\textbf{84.32} &69.76 &74.46 &65.79 &81.27 &75.66 &82.44\\
mAUC$^{\textit{total}}$ &\textbf{79.73} &-  & 75.08 &-  & 75.17 &-  &-  & 82.58  &- &73.67 &- & 82.62 & 75.54 &\textbf{82.65} \\
\bottomrule
\end{tabular}}
\label{tab:nihxpert}
\vspace{-10pt}
\end{table*}

\begin{table}[t]
\caption{Ablation study on $K$/$\alpha$ with single domain local nodes for chest x-ray classification.  ``Centralized" denotes the result of centralized training with all samples from one dataset. }
\centering
\resizebox{\columnwidth}{!}
{
\begin{tabular}
{c|c|c|c|cc|cc}
\hline
&\textcolor{black}{Attention} &\textcolor{black}{Dropout} & \multirow{2}{*}{Centralized} &\multicolumn{2}{c|}{$\alpha=1$} &\multicolumn{2}{c}{$\alpha=0.1$}\\
&\textcolor{black}{Bound} &\textcolor{black}{Rate} & &$K=3$ &$K=5$  &$K=3$ &$K=5$ \\\hline
\multirow{2}{*}{NIH}  &\textcolor{black}{\xmark} &\textcolor{black}{0} &\multirow{4}{*}{\textbf{79.73}} &\textcolor{black}{73.94}  &\textcolor{black}{73.39} &\textcolor{black}{64.26} &\textcolor{black}{66.89}\\ 
&\textcolor{black}{\cmark} &\textcolor{black}{0} & &75.17 &75.12 &67.08 &69.30 \\
\multirow{2}{*}{CXR14}  &\textcolor{black}{\cmark} &\textcolor{black}{$\frac{1}{K}$} & &\textcolor{black}{73.72} &\textcolor{black}{74.01} &\textcolor{black}{64.81} &\textcolor{black}{67.95}\\
&\textcolor{black}{\cmark} &\textcolor{black}{$\frac{2}{K}$} & &\textcolor{black}{69.87} &\textcolor{black}{73.64} &\textcolor{black}{60.93} &\textcolor{black}{65.88}\\ \hline
\multirow{4}{*}{CheXpert} &\textcolor{black}{\xmark} &\textcolor{black}{0} &\multirow{4}{*}{82.58} &\textcolor{black}{81.31}  &\textcolor{black}{80.76} &\textcolor{black}{75.09} &\textcolor{black}{76.86}  \\ 
&\textcolor{black}{\cmark} &\textcolor{black}{0} & &\textbf{82.62}  &82.71 &77.23 &79.50 \\
&\textcolor{black}{\cmark} &\textcolor{black}{$\frac{1}{K}$} & &\textcolor{black}{80.78} &\textcolor{black}{81.52} &\textcolor{black}{74.42} &\textcolor{black}{76.17}\\
&\textcolor{black}{\cmark} &\textcolor{black}{$\frac{2}{K}$} & &\textcolor{black}{78.25} &\textcolor{black}{78.65} &\textcolor{black}{72.10} &\textcolor{black}{74.61}\\
\hline
\end{tabular}}
\label{tab:cxrab1}
\vspace{-10pt}
\end{table}

\subsection{Cross-domain Analysis}
Our method maintains generalizability while distilling knowledge from multiple locals with cross-domain public data. Built upon the theories from domain adaptation~\cite{ben2010theory,blitzer2008learning,planche2020bridging}, this section gives a theoretical analysis with a performance bound for the aggregated central model.

We suppose the input space is denoted by $\mathcal{X}$, and the source and target domain are represented as $\mathcal{D}^S$ and $\mathcal{D}^T$, respectively.
Given $h$ as the hypothesis function and $g$ as the ground-truth labeling function, we can infer the error as $\epsilon_{\mathcal{D}^S}(h,g)= \mathbb{E}_{x \sim \mathcal{D}^S}[|h(x)-g(x)|]$, where $\epsilon_{\mathcal{D}^S}$ and $\epsilon_{\mathcal{D}^T}$ represent the risk of $h$ on $\mathcal{D}^S$ and $\mathcal{D}^T$. To evaluate the distance between two domain distributions $\mathcal{U}$, $\mathcal{U'}$ on the hypothesis space $\mathcal{H}$, \cite{ben2010theory} introduces $\mathcal{H}$-divergence $d_{\mathcal{H}}(\mathcal{U}, \mathcal{U}') = 2\operatorname{sup}_{A \in \mathcal{A}_{\mathcal{H}}}|\operatorname{Pr}_{\mathcal{D}}(A) - \operatorname{Pr}_{\mathcal{D}'}(A)|$, where $\mathcal{A}_{\mathcal{H}}$ denotes a collection of subsets of $\mathcal{X}$ which support the hypothesis in $\mathcal{H}$. 
The symmetric different space is defined as $\mathcal{H} \Delta \mathcal{H}= \{ h(x) \bigoplus h'(x)| h, h' \in \mathcal{H}\}$ ($\bigoplus$ represents the XOR operation).  Then we have the following theorem for the generalizability between two domains \cite{blitzer2008learning}: \\
\textbf{Theorem 1. Generalization bounds.}
\textit{Let $\mathcal{H}$ be a hypothesis space with VC dimension $d$, $\mathcal{U}^S$ and $\mathcal{U}^T$ each be unlabeled samples of size $N$, drawn from $\mathcal{D}^S$ and $\mathcal{D}^T$ respectively.  For any $h \in \mathcal{H}$ and $\delta \in (0,1)$, the following holds with probability at least $1-\delta$ (over the choice of the samples):}
\begin{equation}
\label{eq:genbound}
\begin{aligned}
     \epsilon_{\mathcal{D}^T}(h) \leq & \epsilon_{\mathcal{D}^S}(h) + \frac{1}{2} {d}_{\mathcal{H} \Delta \mathcal{H}} (\mathcal{U}^S, \mathcal{U}^T) \\
     &+ 4 \sqrt{\frac{2d\operatorname{log}(2N)+\operatorname{log}(\frac{2}{\delta})}{N}} + \lambda,
\end{aligned}
\end{equation}
\textit{where $\lambda=\epsilon_{\mathcal{D}^S}(h^*)+\epsilon_{\mathcal{D}^T}(h^*)$ and $h^*$ is the ideal joint hypothesis minimizing the combined error: $h^* = \operatorname{argmin}_{h \in \mathcal{H}} \epsilon_{\mathcal{D}^S}(h^*)+\epsilon_{\mathcal{D}^T}(h^*)$.}

In our case, $\mathcal{D}^S$ is the domain of private data across $K$ local nodes $\mathcal{D}^S = \{ \mathcal{D}^k \}$, and $\mathcal{D}^T$ = $\mathcal{D}^0$ is the domain of public data, where we assume $|\mathcal{D}^0|=N$ and $\sum_{k} |\mathcal{D}^k|=N$. 
Given a local model $h_{\mathcal{D}^k}$ trained on data $\mathcal{D}^k$, we learn a central model $h_{\mathcal{D}^0}$ with unlabeled public data ${\mathcal{D}^0}$ through weighted aggregation: $h_{\mathcal{D}^0} = \sum_{k} \omega_k h_{\mathcal{D}^k}$,  where $\sum_{k} \omega_k=1$. 
As proved in \cite{peng2020federated}, the overall private data is $\mathcal{U}^S = \sum_{k} \omega_k \mathcal{U}^k$, and ${d}_{\mathcal{H} \Delta \mathcal{H}} (\sum_{k} \omega_k \mathcal{U}^k, \mathcal{U}^0) \leq \sum_{k} \omega_k( \frac{1}{2} {d}_{\mathcal{H} \Delta \mathcal{H}} (\mathcal{U}^k, \mathcal{U}^0) )$. We then rewrite Eq. \ref{eq:genbound} and have the weighted generalization bound as Eq. \ref{eq:genbound2}, where we note that the test error of central model $\epsilon_{\mathcal{D}^0}$ is bounded by that of local model $\epsilon_{\mathcal{D}^k}$, the domain gap between local data and public data ${d}_{\mathcal{H} \Delta \mathcal{H}} (\mathcal{U}^k, \mathcal{U}^0)$, the function VC dimension $d$, and the data size $N$. 
\begin{equation}
\label{eq:genbound2}
\begin{aligned}
\epsilon_{\mathcal{D}^0}(h_{\mathcal{D}^0}) 
&\leq \epsilon_{\mathcal{D}^k}(\sum_{k }{\omega_k h_{\mathcal{D}^k}}) + \frac{1}{2} {d}_{\mathcal{H} \Delta \mathcal{H}} (\sum_{k} \omega_k \mathcal{U}^k, \mathcal{U}^0) \\
& \quad + 4 \sqrt{\frac{2d\operatorname{log}(2N)+\operatorname{log}(\frac{2}{\delta})}{N}} + \lambda_\omega \\
&\leq \epsilon_{\mathcal{D}^k} (\sum_{k }{\omega_k h_{\mathcal{D}^k})} + \sum_{k} \omega_k\left( \frac{1}{2} {d}_{\mathcal{H} \Delta \mathcal{H}} (\mathcal{U}^k, \mathcal{U}^0) \right) \\ 
& \quad + 4 \sqrt{\frac{2d\operatorname{log}(2N)+\operatorname{log}(\frac{2}{\delta})}{N}} + \lambda_\omega.
\end{aligned}
\end{equation}


\section{Experiments}
\label{sec:experiment}
We employ one-shot distillation (each local model transfers its prediction over public data only once, and these local products are used for numerous steps of central training) for bandwidth-sensitive tasks like segmentation and reconstruction \textcolor{black}{for} communication efficiency. 
\subsection{Chest X-Ray Image Classification}
\subsubsection{Datasets}
\label{4.1.1}
We evaluate our method on a multi-label classification task with standard chest-x-ray datasets: NIH chestX-ray14 (NIH CXR14)~\cite{wang2017chestx} and CheXpert~\cite{irvin2019chexpert} \textcolor{black}{as locally held private data}. NIH CXR14 consists of 112,120 frontal-view x-ray images scanned from 32,717 patients labeled with 14 diseases. CheXpert contains 224,316 chest radiographs from 65,240 patients labeled with 14 common chest radiographs,  including both frontal-view and lateral-view.  For public data, we use 26,684 x-ray images in the RSNA Pneumonia Detection Challenge public data \cite{rsna2018} without using their labels.

\subsubsection{Implementation}
We use NIH CXR14 and CheXpert as domains where private data comes from. 
For samples with multiple positive labels, we randomly choose one and split the dataset across locals using the Dirichlet distribution as in most FL works \cite{hsu2019measuring}, in which the value of $\alpha$ controls the degree of non-IID-ness. A smaller $\alpha$ indicates higher non-IID-ness.
For the total of $K$ local nodes, each dataset is distributed to $K_d=K/2$ local nodes under the cross-domain scenario. \textcolor{black}{Following the validation strategy in \cite{ye2020weakly},} \textcolor{black}{for both datasets} we \textcolor{black}{randomly} sample a fraction (10\%) of the training data  to form the validation set.
For training, we use ResNet-34 with a batch size of 32 and the same data augmentation methods as in \cite{ye2020weakly}. Each local model is trained individually with SGD and CosineAnnealing \cite{loshchilov2016sgdr} and a decreasing learning rate from 1e-3 to 1e-6 across 15 epochs. We use SGD and CosineAnnealing for distillation and a decreasing learning rate from 1e-2 to 1e-3 across 20 epochs. 
\begin{table}
\vspace{-10pt}
\caption{Ablation study on $|\mathcal{D}_0|$ ($K=3$, $\alpha=1$) using CheXpert as local datasets.}
\centering
\scriptsize
{
\begin{tabular}
{c|cccccc}
\toprule
&\multicolumn{6}{c}{\# samples in $\mathcal{D}_0$} \\\cmidrule{2-7}
{Centralized}&1000 &5000 &10000 &15000 &20000 &26684 \\\midrule
82.58 &81.08 &81.23 &81.36 &81.57 &82.37 &82.62 \\
\bottomrule
\end{tabular}}
\label{tab:cxrab2}
\vspace{-10pt}
\end{table} 

\begin{table}
\caption{\textcolor{black}{Comparison with parameter based and distillation based FL methods on chest-x-ray image classification task using CheXpert as local datasets ($K=3$, $\alpha=1$).  ``Centralized" denotes the result of centralized training with all samples from one dataset.}}
\centering
\resizebox{\columnwidth}{!}{
\begin{tabular}
{c|c|cccc}
\hline
&{\textcolor{black}{Centralized}} &\textcolor{black}{FedAvg \cite{mcmahan2017communication}} &\textcolor{black}{FedDF \cite{lin2020ensemble}} &\textcolor{black}{FedMD \cite{li2019fedmd}} &\textcolor{black}{Ours} \\\hline
\textcolor{black}{Distillation} &\textcolor{black}{-} &\textcolor{black}{N} &\textcolor{black}{Y} &\textcolor{black}{Y} &\textcolor{black}{Y}\\
\textcolor{black}{Param. Trans.} &\textcolor{black}{-} &\textcolor{black}{Y} &\textcolor{black}{Y} &\textcolor{black}{N} &\textcolor{black}{N}\\
\textcolor{black}{Privacy} &\textcolor{black}{-} &\textcolor{black}{\xmark} &\textcolor{black}{\xmark} &\textcolor{black}{\cmark} &\textcolor{black}{\cmark} \\
\textcolor{black}{Asynchronous} &\textcolor{black}{-} &\textcolor{black}{\xmark} &\textcolor{black}{\xmark} &\textcolor{black}{\xmark}  &\textcolor{black}{\cmark} \\
\textcolor{black}{mAUC(\%)} &\textcolor{black}{82.58} &\textcolor{black}{79.03} &\textcolor{black}{82.94}  &\textcolor{black}{77.66} &\textcolor{black}{82.62} \\
\hline
\end{tabular}}
\label{tab:cxrcomp}
\vspace{-10pt}
\end{table} 

\subsubsection{Results}
We first study the cases when local samples are from one dataset.
Table~\ref{tab:cxrab1} shows results \wrt when local data are within the domain with a varying number of locals $K$ and non-IID-ness $\alpha$. One can note that our method outperforms centralized training with all local data on CheXpert \textcolor{black}{when $K$=3 and $\alpha$=1}. 
Table ~\ref{tab:cxrab2} shows training results with varying public dataset sizes. The results suggest that, although unlabeled, a more extensive public dataset improves performance. 
\textcolor{black}{To compare with the SOTA FL methods, Table~\ref{tab:cxrcomp} shows whether the method in comparison transfers parameters or employs distillation and analyzes each privacy guarantee and synchronization requirements. We can see that our method outperforms the counterparts with the best utility-privacy trade-off.}

With both datasets \textcolor{black}{as private data}, we conduct cross-domain, cross-site evaluations with FedAD. We distribute the training datasets to $K$=6 local nodes ($K_d$=3 for each dataset). Table~\ref{tab:nihxpert} shows the results of FedAD on cross-domain datasets with $\alpha$=1.
On  CheXpert,  both single domain and cross-domain distillation achieve better performance than centralized training: cross-domain learning  with  FedAD  obtains the best mAUC of 82.65\%,  slightly better than FedAD training with only data from CheXpert (82.62\%).
On NIH CXR14, while training with all data centrally yields the best result (79.73\%), cross-domain trained FedAD still obtains better performance (75.54\%) compared to FedAD trained with single domain data (75.17\%) and compared to the average AUC of local nodes (75.08\%). Table~\ref{tab:nihxpert} also reports results on the six overlapping classes in each test set. It can be seen that FedAD trained with cross-domain data obtains superior performance compared to the two FedAD models trained with single domain data on both test datasets. We note that the FedAD model trained with cross-domain data can classify the 14 classes in total (12 classes from CheXpert and the 8 classes from NIH CXR14), whereas other models can only classify 8 or 12 classes. 

\begin{table}
\vspace{-10pt}
\centering
\caption{Comparisons on BraTS in terms of average Dice score over voxel-level annotations of “whole tumour”, “tumour core”, “enhancing tumour”, and communication efficiency attributes.}
\resizebox{\columnwidth}{!}{
\begin{tabular}
{c|c|cc|c|c}
\hline
\multirow{2}{*}{Method} &Average  &\multicolumn{2}{c|}{Communication Efficiency} &\textcolor{black}{\multirow{2}{*}{Transmit}} &\textcolor{black}{{Privacy}}\\ 
&Dice(\%)$\uparrow$ &Bandwidth(GB)$\downarrow$  &\textcolor{black}{Asynchronous} & &\textcolor{black}{Cost/Risk$\downarrow$}
\\\hline
\multirow{3}{*}{\textcolor{black}{Li \etal \cite{li2019privacy}}} &\textcolor{black}{84.33} &\multirow{3}{*}{\textcolor{black}{64.37}} &\multirow{3}{*}{\textcolor{black}{\xmark}} &\multirow{3}{*}{\textcolor{black}{Parameter}} &\textcolor{black}{$\infty$}\\ 
&\textcolor{black}{81.28} & & & &\textcolor{black}{$\epsilon_1$=1, $\epsilon_3$=1}\\ 
&\textcolor{black}{80.01} & & & &\textcolor{black}{$\epsilon_1$=1, $\epsilon_3$=0.01}\\
\hline
FedMD \cite{li2019fedmd} &75.71 &2154.84 &\textcolor{black}{\xmark} &\textcolor{black}{Distillation} &\textcolor{black}{0}\\
Ours &77.85 &13.36  &\textcolor{black}{\cmark} &\textcolor{black}{Distillation} &\textcolor{black}{0}\\
\textcolor{black}{Standalone}  &\textcolor{black}{73.38$\pm$ 3.44} &\textcolor{black}{-} &\textcolor{black}{-} &\textcolor{black}{-} &\textcolor{black}{-} \\\hline
\end{tabular}
}
\label{tab:brats}
\end{table}

\begin{table}
\vspace{-10pt}
\centering
\caption{Ablation study on output ensemble scheme, attention lower/upper bound, and the modality of public data. T1-weighted images from B, F, I training set are used as local data, and T1-weighted images from B, F, I testing set are used as evaluation.} 
\resizebox{\columnwidth}{!}
{
\begin{tabular}{c|c|c|c|c|c|c}
\hline
\textit{Ensemble scheme} &\cite{lin2020ensemble, hsu2020federated} &Eq.\ref{eqweight} &Eq.\ref{eqweight} &Eq.\ref{eqweight} &Eq.\ref{eqweight} &Eq.\ref{eqweight}\\
\textit{Non-local module} (Eq.\ref{eq:nlocal}) &\xmark &\xmark &\cmark &\cmark &\cmark &\cmark\\
\textit{Attention lower bound} (Eq.\ref{eq:lbound}) &\xmark &\xmark &\xmark &\cmark &\cmark &\cmark\\
\textit{Attention upper bound} (Eq.\ref{eq:ubound}) &\xmark &\xmark &\xmark &\xmark &\cmark &\cmark\\
\textit{Unlabeled public data $\mathcal{D}_0$} &T1w &T1w &T1w &T1w &T1w &T2w \\ \hline
SSIM $\uparrow$ &0.8892 &0.9097 &0.9108 &0.9147  &\textbf{0.9161} &0.9112\\
PSNR $\uparrow$ &32.91 &33.20 &33.24 &33.30  &\textbf{33.38} &33.05\\
\hline
\end{tabular}}
\label{tab:mriab}
\vspace{-10pt}
\end{table}

\subsection{3D Brain Tumor Segmentation}
\subsubsection{Dataset} The BraTS 2018 dataset~\cite{bakas2018identifying} contains multi-parametric pre-operative MRI scans of 285 subjects with brain tumors. Each subject was scanned under the T1-weighted, T1-weighted with contrast enhancement, T2-weighted, and T2 fluid-attenuated inversion recovery (T2-FLAIR) modalities.

\subsubsection{Implementation} 
Following the same protocol as in~\cite{li2019privacy}, we use 242 subjects for the training set and 43 subjects for held-out test set. According to the originating institution, the training set is stratified into three federated local clients. We use the unlabeled validation set of the BraTS 2020 dataset~\cite{menze2014multimodal} as the public data comprising 125 subjects, independent of either private dataset. 
We use the same model structure as~\cite{li2019privacy, myronenko20183d} but with half its channel numbers.  \textcolor{black}{We only use its segmentation branch (no reconstruction branch).} The training strategy is the same as~\cite{myronenko20183d}. The local training takes 20,000 iterations, and local-to-central distillation takes 5,000 iterations. 


\subsubsection{Results} Table~\ref{tab:brats} compares the segmentation performance on BraTS with \textcolor{black}{the SOTA parameter based method \cite{li2019privacy}} and \textcolor{black}{the SOTA distillation based method} \cite{li2019fedmd}. \textcolor{black}{Note we report both naive (non-private) and noisy (less-private) version of \cite{li2019privacy}. The comparison shows that our method achieves the best utility-privacy trade-off. } We can \textcolor{black}{also} observe that \textcolor{black}{when compared with \cite{li2019fedmd},} our method achieves better results with much more efficiency, \ie, lower communication bandwidth and no local synchronization requirement at the same time.

\begin{table}
\vspace{-10pt}
\centering
\caption{Results on in-domain testing sets for MRI image reconstruction. ``Standalone" denotes the locally trained model with individual private data. Under the FL setting, we compare our method with FL-MRCM \cite{Guo_2021_CVPR} quantitatively in terms of SSIM, PSNR, and communication Bandwidth.
} 
\resizebox{\columnwidth}{!}
{
\begin{tabular}{ccc|cc|cccc|c}
\toprule
&\multirow{3}{*}{\rotatebox[origin=c]{90}{{Privacy}}} &\multirow{3}{*}{\rotatebox[origin=c]{90}{{Flexibility}}} &\multicolumn{2}{c|}{\multirow{2}{*}{Data} } &\multicolumn{2}{c}{\multirow{2}{*}{{T1-weighted}}} &\multicolumn{2}{c|}{\multirow{2}{*}{{T2-weighted}}} &\multirow{2}{*}{Bandwidth} \\
& & & & & & & & &\\
& & &\multirow{2}{*}{Train} &\multirow{2}{*}{Test} &\multirow{2}{*}{SSIM$\uparrow$} &\multirow{2}{*}{PSNR$\uparrow$} &\multirow{2}{*}{SSIM$\uparrow$} &\multirow{2}{*}{PSNR$\uparrow$} &\multirow{2}{*}{(GB)$\downarrow$} \\
& & & & & & & & &\\
\midrule
\parbox[t]{1mm}{\multirow{9}{*}{\rotatebox[origin=c]{90}{\shortstack[c]{Standalone}}}} &\multirow{9}{*}{-} &\multirow{9}{*}{-} &\multirow{3}{*}{B} &\cellcolor{gray0}B &\cellcolor{gray0}0.9743 &\cellcolor{gray0}38.81 &\cellcolor{gray0}0.9694 &\cellcolor{gray0}36.53 &\multirow{9}{*}{-}   \\
& & & &\cellcolor{gray1}F &\cellcolor{gray1}0.7787 &\cellcolor{gray1}29.16 &\cellcolor{gray1}0.8028 &\cellcolor{gray1}27.35 & \\
& & & & \cellcolor{gray2}I & \cellcolor{gray2} 0.8948 & \cellcolor{gray2}31.02 & \cellcolor{gray2}0.7692 & \cellcolor{gray2} 27.47 & \\\cmidrule{4-9}
 & & &\multirow{3}{*}{F} &\cellcolor{gray0}B &\cellcolor{gray0}0.9125 &\cellcolor{gray0}33.82 &\cellcolor{gray0}0.9250 &\cellcolor{gray0}33.96 &  \\
 & & & &\cellcolor{gray1}F &\cellcolor{gray1}0.9360 &\cellcolor{gray1}33.83 &\cellcolor{gray1}0.9522 &\cellcolor{gray1}33.84 & \\
 & & & &\cellcolor{gray2}I &\cellcolor{gray2}0.9146 &\cellcolor{gray2}31.26 &\cellcolor{gray2}0.9003 &\cellcolor{gray2}30.67 & \\\cmidrule{4-9}
 & & &\multirow{3}{*}{I} &\cellcolor{gray0}B &\cellcolor{gray0}0.9421 &\cellcolor{gray0}34.98 &\cellcolor{gray0}0.9111 &\cellcolor{gray0}31.76 &  \\
 & & & &\cellcolor{gray1}F &\cellcolor{gray1}0.8919 &\cellcolor{gray1}31.80 &\cellcolor{gray1}0.9092 &\cellcolor{gray1}30.61 & \\
 & & & &\cellcolor{gray2}I &\cellcolor{gray2}0.9615 &\cellcolor{gray2}34.91 &\cellcolor{gray2}0.9336 &\cellcolor{gray2}32.14 & \\\midrule  \midrule

\parbox[t]{1mm}{\multirow{3}{*}{\rotatebox[origin=c]{90}{\shortstack[c]{Central\\-ize}}}} &\multirow{3}{*}{-} &\multirow{3}{*}{-} &\multirow{3}{*}{B,F,I} &\cellcolor{gray0}B &\cellcolor{gray0}0.9557 &\cellcolor{gray0}37.19 &\cellcolor{gray0}0.9398 &\cellcolor{gray0}34.08 &\multirow{3}{*}{-}   \\
& &  & &\cellcolor{gray1}F &\cellcolor{gray1}0.9335 &\cellcolor{gray1}34.58 &\cellcolor{gray1}0.9002 &\cellcolor{gray1}30.47 & \\
& & & &\cellcolor{gray2}I &\cellcolor{gray2}0.9451 &\cellcolor{gray2}33.55 &\cellcolor{gray2}0.8873 &\cellcolor{gray2}30.95 & \\\midrule\midrule

\parbox[t]{1mm}{\multirow{4}{*}{\rotatebox[origin=c]{90}{\shortstack[c]{\cite{Guo_2021_CVPR}}}}}
&\multirow{3}{*}{\xmarkcolor} &\multirow{3}{*}{\rotatebox[origin=c]{90}{\shortstack[c]{Online}}} &\multirow{3}{*}{B,F,I} &\cellcolor{gray0}B &\cellcolor{gray0}\textbf{0.9577} &\cellcolor{gray0}\textbf{36.88} &\cellcolor{gray0}\textbf{0.9308} &\cellcolor{gray0}\textbf{34.28} &\multirow{3}{*}{868.8}  \\
& & &  &\cellcolor{gray1}F &\cellcolor{gray1}0.9023 &\cellcolor{gray1}\textbf{33.63} &\cellcolor{gray1}0.8974 &\cellcolor{gray1}31.24 & \\
& & & &\cellcolor{gray2}I &\cellcolor{gray2}\textbf{0.9362} &\cellcolor{gray2}\textbf{33.29} &\cellcolor{gray2}0.8778 &\cellcolor{gray2}30.44 & \\
\midrule

\parbox[t]{1mm}{\multirow{4}{*}{\rotatebox[origin=c]{90}{\shortstack[c]{Ours}}}}
&\multirow{3}{*}{\cmarkcolor} &\multirow{3}{*}{\rotatebox[origin=c]{90}{\shortstack[c]{Offline}}} &\multirow{3}{*}{B,F,I} &\cellcolor{gray0}B &\cellcolor{gray0}0.9111 &\cellcolor{gray0}34.55 &\cellcolor{gray0}0.9199 &\cellcolor{gray0}34.06 &\multirow{3}{*}{\textbf{866.0}} \\
& & & &\cellcolor{gray1}F &\cellcolor{gray1}\textbf{0.9182} &\cellcolor{gray1}33.37 &\cellcolor{gray1}\textbf{0.9374} &\cellcolor{gray1}\textbf{32.76} & \\
& & & &\cellcolor{gray2}I &\cellcolor{gray2}0.9173 &\cellcolor{gray2}31.72 &\cellcolor{gray2}\textbf{0.9058} &\cellcolor{gray2}\textbf{30.93} & \\
\bottomrule
\end{tabular}}
\vspace{-10pt}
\label{tab:mriin}
\end{table}

\subsection{Brain Magnetic Resonance Image Reconstruction}
\subsubsection{Datasets}
Following the prior art \cite{Guo_2021_CVPR},
we use fastMRI \cite{zbontar2018fastMRI}, IXI \cite{ixi}, BraTS\cite{hdtd-5j88-20} as private data distributed across local nodes and evaluate the corresponding test sets (same data split as \cite{Guo_2021_CVPR}). Guo \etal \cite{Guo_2021_CVPR} reports results with four brain MRI datasets, of which HPKS \cite{jiang2019identifying} is privately held and not available for use at the moment; so we experiment with the remaining publicly available sets as local data: fastMRI, IXI, and BraTS. Besides, we use OASIS-3 \cite{lamontagne2019oasis} as a public dataset.

\noindent\textbf{FastMRI} \cite{zbontar2018fastMRI} (abbreviated as F):  for fastMRI we use T1-weighted images from 3,443 subjects. Of the 3,443, data from 2,583 subjects are used for training, and 860 are used for testing. Besides, we use T2-weighted images from 3,832 subjects, of which 2,874 subjects are for training, and the rest 958 subjects are used for testing. Each subject consists of approximately 15 axial cross-sectional images of brain tissues.

\noindent\textbf{BraTS} \cite{hdtd-5j88-20} (abbreviated as B): BraTS2020 is composed of 494 subjects available for both  T1 and T2-weighted modalities. Of these, 369 subjects are used for training and 125 subjects for testing. Each subject includes approximately 120 axial cross-sectional images of brain tissues for both modalities.

\noindent\textbf{IXI} \cite{ixi} (abbreviated as I): 
IXI dataset has 581 subjects available for the T1-weighted modality, among which 436, 55, and 90 subjects are used for training, validation, and testing respectively. 
For the T2-weighted modality, there are 578 subjects, of which data from 434 subjects are for training, 55 for validation, and  89 for testing. Approximately 150 and 130 axial cross-sectional images of brain tissues for T1 and T2-weighted, respectively, are provided for each subject. 

\noindent\textbf{OASIS-3} \cite{lamontagne2019oasis}: Open Access Series of Imaging Studies (OASIS-3) is a multi-modal dataset. It contains 3,388 subjects for T1w and 3,598 subjects for T2w. All sessions are collected with a 16-channel head coil on 1.5T scanners.  

\begin{figure}
\vspace{-10pt}
\begin{center}
\includegraphics[width=\linewidth]{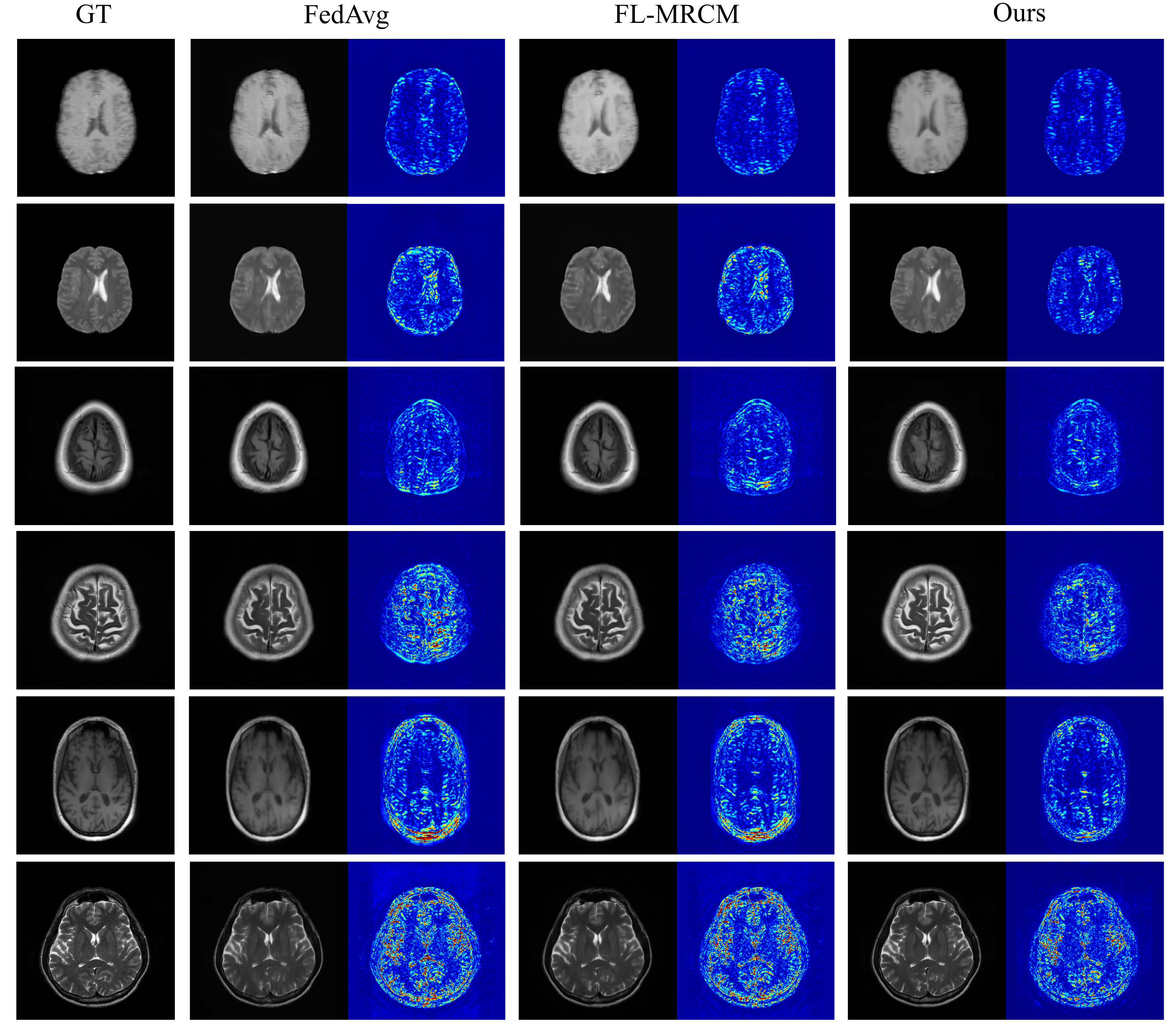}
\caption{Qualitative results of different methods when training with T1/T2-weighted B, F, I as local data and testing on T1 B, T2 B, T1 F, T2 F, T1 I, T2 I test set respectively. The second column of each sub-figure is the error map (absolute difference) between the reconstructed images and the ground truth.}
\label{fig:reconvis}
\end{center}
\vspace{-15pt}
\end{figure}

\begin{table*}
\vspace{-15pt}
\centering
\caption{Results on cross-domain testing sets for MRI image reconstruction.} 
\resizebox{\textwidth}{!}
{
\begin{tabular}{c|c|c|cc|ccc|cc|ccc|cc}
\toprule
\multirow{3}{*}{Method} &\multirow{3}{*}{\textcolor{black}{Transmit}} &\multirow{3}{*}{Privacy} &\multicolumn{2}{c|}{Data}  &\multicolumn{5}{c|}{T1-weighted}  &\multicolumn{5}{c}{T2-weighted}\\\cmidrule{4-15}
& & &\multirow{2}{*}{Train} &\multirow{2}{*}{Test} &\multirow{2}{*}{SSIM$\uparrow$} &\multirow{2}{*}{PSNR$\uparrow$} &\multirow{2}{*}{\# test subjects} &\multicolumn{2}{c|}{\textit{wAverage}}  &\multirow{2}{*}{SSIM$\uparrow$} &\multirow{2}{*}{PSNR$\uparrow$} &\multirow{2}{*}{\# test subjects}  &\multicolumn{2}{c}{\textit{wAverage}}\\
& & & & & & & &SSIM$\uparrow$ &PSNR$\uparrow$  & & & &SSIM$\uparrow$ &PSNR$\uparrow$ \\\midrule
\multirow{3}{*}{FedAvg\cite{mcmahan2017communication}} &\multirow{3}{*}{\textcolor{black}{Parameter}} &\multirow{3}{*}{\xmarkcolor} &B,F &\cellcolor{gray0}I &\cellcolor{gray0}0.9086 &\cellcolor{gray0}31.46 &\cellcolor{gray0}90 &\multirow{3}{*}{0.8533}  &\multirow{3}{*}{31.49}  &\cellcolor{gray0}0.8532 &\cellcolor{gray0}29.42 &\cellcolor{gray0}89 &\multirow{3}{*}{0.8608} &\multirow{3}{*}{30.36} \\
& & &F,I &\cellcolor{gray1}B &\cellcolor{gray1}0.9268 &\cellcolor{gray1}34.69 &\cellcolor{gray1}125 & & &\cellcolor{gray1}0.9062 &\cellcolor{gray1}33.11 &\cellcolor{gray1}128 & & \\
& & &B,I &\cellcolor{gray2}F &\cellcolor{gray2}0.8369 &\cellcolor{gray2}31.03 &\cellcolor{gray2}860 & & &\cellcolor{gray2}0.8556 &\cellcolor{gray2}30.09 &\cellcolor{gray2}958 & & \\
\midrule

\multirow{3}{*}{FL-MRCM\cite{Guo_2021_CVPR}} &\multirow{3}{*}{\textcolor{black}{Parameter}} &\multirow{3}{*}{\xmarkcolor} &B,F &\cellcolor{gray0}I &\cellcolor{gray0}0.9157 &\cellcolor{gray0}31.74 &\cellcolor{gray0}90 &\multirow{3}{*}{\textbf{0.8553}} &\multirow{3}{*}{\textbf{32.39}} &\cellcolor{gray0}0.8354 &\cellcolor{gray0}29.28 &\cellcolor{gray0}89 &\multirow{3}{*}{0.8632} &\multirow{3}{*}{30.00} \\
& & &F,I &\cellcolor{gray1}B &\cellcolor{gray1}0.9486 &\cellcolor{gray1}35.78 &\cellcolor{gray1}125 & & &\cellcolor{gray1}0.9041 &\cellcolor{gray1}33.15 &\cellcolor{gray1}128 \\
& & &B,I &\cellcolor{gray2}F &\cellcolor{gray2}0.8354 &\cellcolor{gray2}31.96 &\cellcolor{gray2}860 & & &\cellcolor{gray2}0.8604 &\cellcolor{gray2}29.66 &\cellcolor{gray2}958 \\
\midrule

\multirow{3}{*}{\textcolor{black}{FedDF\cite{lin2020ensemble}}} &{\textcolor{black}{Parameter}} &\multirow{3}{*}{{\xmarkcolor}} &\textcolor{black}{B,F} &\cellcolor{gray0}\textcolor{black}{I} &\cellcolor{gray0}\textcolor{black}{0.9209} &\cellcolor{gray0}\textcolor{black}{32.11} &\cellcolor{gray0}\textcolor{black}{90} &\multirow{3}{*}{\textcolor{black}{0.8665}} &\multirow{3}{*}{\textcolor{black}{32.75}} &\cellcolor{gray0}\textcolor{black}{0.8781} &\cellcolor{gray0}\textcolor{black}{30.20} &\cellcolor{gray0}\textcolor{black}{89} &\multirow{3}{*}{\textcolor{black}{0.8819}} &\multirow{3}{*}{\textcolor{black}{30.77}} \\
&{\textcolor{black}{$+$}} & &\textcolor{black}{F,I} &\cellcolor{gray1}\textcolor{black}{B} &\cellcolor{gray1}\textcolor{black}{0.9561} &\cellcolor{gray1}\textcolor{black}{35.92} &\cellcolor{gray1}\textcolor{black}{125} & & &\cellcolor{gray1}\textcolor{black}{0.9154} &\cellcolor{gray1}\textcolor{black}{33.96} &\cellcolor{gray1}\textcolor{black}{128} \\
&{\textcolor{black}{Distillation}} & &\textcolor{black}{B,I} &\cellcolor{gray2}\textcolor{black}{F} &\cellcolor{gray2}\textcolor{black}{0.8479} &\cellcolor{gray2}\textcolor{black}{32.36} &\cellcolor{gray2}\textcolor{black}{860} & & &\cellcolor{gray2}\textcolor{black}{0.8775} &\cellcolor{gray2}\textcolor{black}{30.37} &\cellcolor{gray2}\textcolor{black}{958} \\
\midrule

\multirow{3}{*}{Ours} &\multirow{3}{*}{\textcolor{black}{Distillation}}&\multirow{3}{*}{\cmarkcolor} &B,F &\cellcolor{gray0}I &\cellcolor{gray0}0.9141 &\cellcolor{gray0}31.26 &\cellcolor{gray0}90 &\multirow{3}{*}{0.8521} &\multirow{3}{*}{31.54} &\cellcolor{gray0}0.8883 &\cellcolor{gray0}29.92 &\cellcolor{gray0}89 &\multirow{3}{*}{\textbf{0.8827}} &\multirow{3}{*}{\textbf{30.59} } \\
& & &F,I &\cellcolor{gray1}B &\cellcolor{gray1}0.9052 &\cellcolor{gray1}33.25 &\cellcolor{gray1}125 & & &\cellcolor{gray1}0.8964 &\cellcolor{gray1}33.24 &\cellcolor{gray1}128 & &\\
& & &B,I &\cellcolor{gray2}F &\cellcolor{gray2}0.8533 &\cellcolor{gray2}31.62 &\cellcolor{gray2}860 & & &\cellcolor{gray2}0.8805 &\cellcolor{gray2}30.31 &\cellcolor{gray2}958 & & \\
\midrule \midrule

\multirow{3}{*}{Centralized} &\multirow{3}{*}{\textcolor{black}{-}} &\multirow{3}{*}{-} &B,F &\cellcolor{gray0}I &\cellcolor{gray0}0.9015 &\cellcolor{gray0}31.03 &\cellcolor{gray0}90 &\multirow{3}{*}{0.8827} &\multirow{3}{*}{32.76}  &\cellcolor{gray0}0.8763 &\cellcolor{gray0}29.22 &\cellcolor{gray0}89 &\multirow{3}{*}{0.8846} &\multirow{3}{*}{30.56} \\
& & &F,I &\cellcolor{gray1}B &\cellcolor{gray1}0.9246 &\cellcolor{gray1}34.75 &\cellcolor{gray1}125 & & &\cellcolor{gray1}0.9045 &\cellcolor{gray1}33.07 &\cellcolor{gray1}128 & & \\
& & &B,I &\cellcolor{gray2}F &\cellcolor{gray2}0.8747 &\cellcolor{gray2}32.65 &\cellcolor{gray2}860 & & &\cellcolor{gray2}0.8827 &\cellcolor{gray2}30.33 &\cellcolor{gray2}958 & & \\
\bottomrule
\end{tabular}}
\label{tab:mricross}
\vspace{-10pt}
\end{table*}

\subsubsection{Implementation}
Following \cite{Guo_2021_CVPR}, we subsample the given k-space by multiplying with a mask, where the acceleration factor (AF) is set as 4.
The 2D MRI images are preprocessed with zero padding and then cropped to the size of $256 \times 256$.
We utilize the same U-Net \cite{ronneberger2015u} style encoder-decoder architecture for the reconstruction networks as the one provided in FL-MRCM \cite{Guo_2021_CVPR}. 
A minor difference in the architecture with \cite{Guo_2021_CVPR} comes from the additional residual non-local block \ref{eq:nlocal} deployed on the bottom features of U-Net before forwarding into a sequence of the up-sampling layer. The size of these features $J \times H \times W$ are $512 \times 16 \times 16$. 
For local training, the network is trained with an Adam optimizer using a constant learning rate of $1e^{-4}$ with 20 epochs.
For distillation, the central model is trained with an RMSprop optimizer using a constant learning rate of $1e^{-4}$ with five epochs in one communication round.

\subsubsection{Results}
We first perform ablation studies on the impact of the output ensemble scheme, the proposed attention distillation bound, and the modality of unlabeled public data. Here we use the T1-weighted images from BraTS (B), fastMRI (F), and IXI (I) as private datasets and unlabeled T1-weighted images from OASIS-3 as public data; and we report results on the aggregated T1-weighted test images from B, F, and I.
From Table \ref{tab:mriab} we can see the superiority of the importance-weighted ensemble beyond the average ensemble typically used in previous FL works \cite{lin2020ensemble, hsu2020federated}. 
Our proposed attention to upper/lower bound constraint further improves the reconstruction performance with higher SSIM and PSNR. In addition, Table \ref{tab:mriab} shows the comparison of using T1-weighted and T2-weighted images \textcolor{black}{of OASIS-3} as public data. The results demonstrate the robustness of our method to public data from a different domain (locally held data and data used for distillation are all from other datasets) and even different modalities (all three local nodes have T1-weighted images for training while public data is T2-weighted).

Second, we compare the performance with the prior arts  \cite{Guo_2021_CVPR} when taking T1/T2-weighted images from B, F, and I as local data and  unlabeled T1/T2-weighted images of OASIS-3 as public data. The results are reported on the corresponding test sets \textcolor{black}{of local data}, respectively. From Table \ref{tab:mriin} we can see that, when compared with the prior art \cite{Guo_2021_CVPR},  our method achieves very competitive reconstruction performance in terms of SSIM and PSNR with higher communication efficiency while  at the same time maintaining the local data privacy by not sharing local model parameters/gradients or any product inferred from local private local data. The counterpart \cite{Guo_2021_CVPR} not only iteratively shares local model parameters but also shares the features inferred on each local private data. 
Besides the superior guarantees \wrt data privacy, 
our method demonstrates  higher communication efficiency through lower bandwidth and higher flexibility (offline communication without any synchronization requirements on the local model). Notably, when testing on T2-weighted F and I, we achieved better performance than centralized training (collecting local data together for training), \eg, on T2-weighted F, we achieved 0.9374/32.76 SSIM/PSNR over the 0.9002/30.47 SSIM/PSNR of centralized training. The reason is that we utilize additional unlabeled, non-sensitive, cross-domain public data, which, we assume, are easily acquired in real-world clinical scenarios. Comparisons of qualitative results are shown in Figure \ref{fig:reconvis}. 

We leverage cross-domain test sets (different domains with the local data) to evaluate the generalizability. 
Table \ref{tab:mricross} \textcolor{black}{compares our method with the SOTA FL methods. It} shows that our privacy-preserving method owns comparable generalizability with the prior arts \cite{mcmahan2017communication, Guo_2021_CVPR, lin2020ensemble}, which share iterative local model parameters \textcolor{black}{and therefore risk privacy leakage}.

\vspace{-5pt}
\section{Conclusions}
\label{sec:conclude}
In this work, we propose a novel distillation-based federated learning framework (FedAD) that can, in principle, preserve local data privacy by using only unlabeled and domain-robust public data. 
To address the communication bottleneck comprehensively, we employ a one-way (offline) knowledge distillation process with an importance-weighted ensemble and attention-bound constraints.
We demonstrate that our proposed attention ensemble scheme can balance the consensus and diversity across locals to handle the inherent heterogeneity in FL scenarios. 
Extensive experiments on various medical image analysis and imaging tasks including classification, segmentation, as well as MR reconstruction  using cross-domain and heterogeneous data distributions highlight the efficacy of FedAD 
and its preservation of local data privacy. 
With privacy being a critical topic for real-world medical applications, we believe our proposed FL framework is able to facilitate 
privacy-abiding learning across various hospital sites \textcolor{black}{and extend to other medical image applications such as object detection and instance segmentation}. 
Future work includes further generalizing the FedAD framework so that it is more task-agnostic, and relaxing \textcolor{black}{or eliminating} the requirement of \textcolor{black}{real} data used in the distillation.

\bibliographystyle{IEEEtran}
\bibliography{bibFile}

\end{document}